\documentclass[review]{elsarticle}

\usepackage{comment}
\usepackage{lineno,hyperref}
\usepackage[ruled,linesnumbered]{algorithm2e}
\usepackage{amsmath,amssymb,amsfonts}
\usepackage{algorithmic}
\usepackage{graphicx}
\usepackage{textcomp}
\usepackage{xcolor}
\usepackage{url}
\usepackage{multirow}
\usepackage{booktabs}
\hypersetup{pdfauthor=author}
\modulolinenumbers[5]

\journal{Journal of Image and Vision Computing}

\begin{document}

\begin{frontmatter}

\title{Knowledge Distillation Methods for Efficient Unsupervised Adaptation Across Multiple Domains}

\author{Le Thanh Nguyen-Meidine\corref{cor1}\fnref{fn1}}
\ead{le-thanh.nguyen-meidine.1@ens.etsmtl.ca}

\author{Atif Belal\fnref{fn3}}
\ead{abelal@myamu.ac.in}

\author{Madhu Kiran\fnref{fn1}}
\ead{madhu.kiran.1@ens.etsmtl.ca}

\author{Jose Dolz\fnref{fn1}}
\ead{jose.dolz@etsmtl.ca}

\author{Louis-Antoine Blais-Morin\fnref{fn2}}
\ead{lablaismorin@genetec.com}

\author{Eric Granger\fnref{fn1}}
\ead{eric.granger@etsmtl.ca}

\address[mysecondaryaddress]{1100 Notre-Dame St W, Montreal, Quebec H3C 1K3, Canada}

\cortext[cor1]{Corresponding author}
\fntext[fn1]{Laboratoire d'imagerie, de vision et d'intelligence artificielle (LIVIA), Ecole de technologie superieure, Montreal, Canada.}
\fntext[fn2]{Genetec Inc., Montreal, Canada.}
\fntext[fn3]{Aligarh Muslim University, Aligarh, India.}

\begin{abstract}
Beyond the complexity of CNNs that require training on large annotated datasets, the domain shift between design and operational data has limited the adoption of CNNs in many real-world applications. For instance, in person re-identification, videos are captured over a distributed set of cameras with non-overlapping viewpoints. The shift between the source (e.g. lab setting) and target (e.g. cameras) domains may lead to a significant decline in recognition accuracy. Additionally, state-of-the-art CNNs may not be suitable for such real-time applications given their computational requirements. Although several techniques have recently been proposed to address domain shift problems through unsupervised domain adaptation (UDA), or to accelerate/compress CNNs through knowledge distillation (KD), we seek to simultaneously adapt and compress CNNs to generalize well across multiple target domains. 
In this paper, we propose a progressive KD approach for unsupervised single-target DA (STDA) and multi-target DA (MTDA) of CNNs. Our method for KD-STDA adapts a CNN to a single target domain by distilling from a larger teacher CNN, trained on both target and source domain data in order to maintain its consistency with a common representation.
This method is extended to address MTDA problems, where multiple teachers are used to distill multiple target domain knowledge to a common student CNN. A different target domain is assigned to each teacher model for UDA, and they alternatively distill their knowledge to the student model to preserve specificity of each target, instead of directly combining the knowledge from each teacher using fusion methods.
Our proposed approach is compared against state-of-the-art methods for compression and STDA of CNNs on the Office31 and ImageClef-DA image classification datasets. It is also compared against state-of-the-art methods for MTDA on Digits, Office31, and OfficeHome. In both settings -- KD-STDA and KD-MTDA -- results indicate that our approach can achieve the highest level of accuracy across target domains, while requiring a comparable or lower CNN complexity. 
\end{abstract}


\end{frontmatter}

\section{Introduction}


Convolutional neural networks (CNNs) have achieved state-of-the-art performance in many visual recognition tasks, i.e., classification, object detection, and segmentation. However, a key limitation that hampers their scalability is their poor generalization to data from new unseen domains, where there exists a shift between data collected for training, source domain, and one/multiple deployment environments, target domains. Particularly, in real-world video surveillance applications like person re-identification, deep Siamese networks are commonly trained to process videos captured over a distributed network of cameras. The variations in data distributions in terms of the different types of the cameras, viewpoints, and capture conditions can cause a significant domain shift, and degrade accuracy \cite{UDA_DISS}. Another drawback of state-of-the-art CNNs is the computational complexity, which impedes its adoption in real-time applications. 
Although it is possible to address the domain shift problem by annotating target domain data and applying supervised transfer learning, the cost of collecting and annotating a large image dataset for each target domain can be prohibitively costly. To alleviate this issue when only unlabeled target data is available, several unsupervised domain adaptation (UDA) techniques have been proposed. Currently, most of UDA techniques focus on learning domain invariant features by minimizing a distance or discrepancy between the two data distributions \cite{MMD_ICLR}, to encourage domain confusion by using adversarial losses \cite{GRL,ADDA}, or both \cite{WD_DA_GAN}. Lastly, it is also possible to learn a mapping between source and target images \cite{DomainMapping1,DomainMapping2}, such that images captured in different domains have a similar appearance, which mimics standard supervised learning. 

While these techniques are very useful for single-target domain adaptation (STDA), they fail to adapt to a multi-target domain adaptation (MTDA) scenario, a relatively unexplored setting of practical importance, where a single model is trained to generalize well across multiple different target domains. A simple solution would be to solve the problem of MTDA by having one CNN trained on each target domain. Nevertheless, this would be very costly in terms of computational resources. For instance, in person re-identification, MTDA can be addressed by adapting a CNN per target domain, or blending target data to adapt a common CNN, although in many real-world applications, these solutions are too costly, and may lead to a reduction in accuracy. Current literature tackles the MTDA task by either taking advantage of domain labels from each target domain \cite{MTDA_Theoric} or ignoring these labels and mix all the target domains together \cite{AMEAN}. The Figure \ref{fig:DAvsMTDA} illustrates the difference between STDA, MTDA and some possible solutions for an MTDA setting.

Current techniques allow to accelerate and compress CNNs while preserving its accuracy. These approaches include: quantization \cite{Quanti1, Quanti2,Quanti3}, low-rank approximation \cite{LRA_Jader, Coordinating}, knowledge distillation (KD) \cite{HIntonKD, RelationalKD, Overhaul, TeachingAssistantKD} and network pruning \cite{Lottery,Molchanov_2019_CVPR, NetworkPruningviaTransformableArchitectureSearch,liu2018rethinking,PruningFPGM}. However, the direction of jointly compressing and adapting CNN to a target domain remains largely unexplored. To the best of our knowledge, only the Transfer Channel Pruning (TCP) technique \cite{TCP} has been proposed to first solve the UDA problem, and then pruning the adapted model. As for using UDA and KD, while recent work \cite{TeachingToAdapt,orbesarteaga2019knowledge} resort to KD to address the UDA task, they focus on reducing the domain gap, rather than the computational cost. Furthermore, most KD techniques rely on labeled dataset whereas in the UDA setting, we do not have access to these labels, which represents an important challenge for the co-joint optimization of KD and UDA.

While it is possible to adopt a compressed CNN to a new target domain, we argue that this scenario will likely degrade the performance of the CNN since over-parametrization is often important for generalization \cite{NIPS2019_8847}. Another alternative could be to first adapt a model to the target domain and then compress it. However, we experimentally demonstrate that such a scenario results in poor performance due to the lack of label information for knowledge distillation. In this paper, we resolve the aforementioned problem by progressively distilling knowledge from a teacher model that is continuously doing domain adaptation to the student. We argue that this can improve the performance since the student learns how to adapt to a new domain instead of learning directly from a previously targeted domain. In addition, we overcome the problem of unsupervised KD for the student model by jointly optimizing UDA and KD and by keeping the student model consistent with the source domain. Our proposed approach can also be generalized for a multi-target domain adaptation -- MTDA, where there exist multiple target domains-- by assigning each teacher to a different target domain. 


In order to address the problem of CNN complexity and domain shift over one or multiple domains, this paper introduces the following contributions. 
1) we propose a new approach for joint KD and UDA that allows training CNNs such that they generalize well on one or multiple target domains;
2) we introduce a consistency loss that ensures that the student models also learn source domain knowledge from the teacher model in order to overcome the challenges of unsupervised KD on target data; 
3) this paper extends our previous work \cite{IJCNN_KD_UDA} in a substantial way by providing new algorithms and extensive experimental results to validate our approach on different UDA and feature distillation techniques; 
4) finally, we extend to the MTDA setting, where our KD-MTDA approach relies on multiple teachers to provide a higher capacity to generalize for multiple target domains, as well to distill their compressed knowledge to a smaller common student CNN that can perform well across multiple target domains.


The rest of this paper is organised as follows. Section 2 provides an overview of state-of-the-art methods for compression and UDA with CNNs. Then, Section 3 introduces our proposed KD approaches for STDA and MTDA of CNNs. The experimental methodology employed for validation, and the results and discussions are described in Sections 4 and 5, respectively.

\section{Related Work}

\subsection{CNN acceleration and compression techniques:}

Time complexity generally depends mostly on the CNN convolutional layers, while memory complexity (number of parameters) depends mostly on the fully connected layers.  
Currently, one of the most popular ways to reduce CNN complexity is to apply pruning techniques \cite{Lottery,Molchanov_2019_CVPR, NetworkPruningviaTransformableArchitectureSearch,liu2018rethinking,PruningFPGM, imavis_fisher_pruning, imavis_pruning_falfnet}, using either weight or filter-level pruning. Another popular strategy to lower the complexity  is quantization \cite{Quanti1, Quanti2, Quanti3}, which reduces the complexity by reducing the representation of weights into a lower precision, i.e., converting 64-bit float precision to an 8-bit integer. Low rank decomposition techniques \cite{LRA_Jader, Coordinating} have also been employed to accelerate CNNs by decomposing weight tensor into lower rank approximation. Lastly, KD \cite{HIntonKD, RelationalKD, Overhaul, TeachingAssistantKD} aims to transfer knowledge from a larger teacher model to a smaller student model, therefore, preserving teacher accuracy on the student, while reducing complexity.

In this paper, we focus on KD based techniques on the intuition that a CNN can efficiently learn how to adapt to a target domain, improving its accuracy on target domain data. Also, we argue that by leveraging over-parametrization \cite{NIPS2019_8847}, our approach can provide a better generalization capacity than when applying UDA on a pruned CNN which has fewer parameters to optimize. Additionally, KD also provides a natural generalization for an MTDA setting where each teacher is responsible for a target domain. Two main approaches have been proposed for distilling a teacher's knowledge to a student -- either based on network outputs \cite{HIntonKD} or on intermediate features \cite{Overhaul}. For example, a temperature-based softmax was used in \cite{HIntonKD} to generate softer versions of the teacher network outputs, facilitating the learning of a student model. In contrast, Heo et al. \cite{Overhaul} enforce similarities between the teacher and student models at intermediate features layers, by minimizing a partial L2 distance. In particular, if the value measures on the student model is smaller than the value of the teacher, and both are negative, then the result is set to 0. A margin ReLU is also employed since the authors argue that negative value from the network is also useful. Another work \cite{TeachingAssistantKD} proposes to solve the problem of the gap between teacher and student by integrating a teaching assistant that serves as an intermediate student that first learns from the teacher then it will distill the information to the student using the output of its network.

\subsection{Single-target domain adaptation:}



Unsupervised domain adaptation (UDA) techniques alleviate the problem of domain shift of a deep learning model that is trained on a labeled source domain data, using unlabeled data from a target domain. Currently, UDA can be achieved by encouraging domain confusion \cite{GRL,ADDA}, learning domain-invariant features \cite{MMD_ICLR, imavis_adv_da_wasserstein}, mapping between source and target domains \cite{DomainMapping_Pixel_Level, ConGAN_DA_MAP, imavis_da_seg}, ensemble learning \cite{Tri_net_DA_Ensemble}, statistic normalization \cite{DA_Normalization_statistics} and target discriminate methods\cite{Wei2018GenerativeAG}. Domain confusion between source and target domain has been done by employing an adversarial loss\cite{ADDA, imavis_adv_da_wasserstein} or by using a gradient reversal layer like \cite{GRL} in combination with a domain classifier. Differently, a model can also learn domain-invariant features, 
for example, by minimizing the Maximum Mean Discrepancy (MMD) between source and target \cite{MMD_ICLR}. Other works such as \cite{DomainMapping_Pixel_Level, ConGAN_DA_MAP} propose to find a mapping from the source to the target domain or vice-versa, typically based on Generative Adversarial Networks (GAN) \cite{DomainMapping_Pixel_Level}. This setting is commonly regarded as a  zero-sum game between two networks, a discriminator and a generator. While the generator tries to fool the discriminator by generating images with target style, the discriminator tries to distinguish the domain the images belong to.  
In \cite{DomainMapping_CADA}, the authors further improve mapping by performing the adaptation at the feature level. Another strategy is to employ multiple models as an ensemble or as self-ensembling (multiple models at different times) such that they can produce reliable pseudo-labels on target domains\cite{Tri_net_DA_Ensemble}. Last, some other researchers have proposed to adapt the batch norm statistics \cite{DA_Normalization_statistics} instead of adapting the network's layers.

\subsection{Multi-target domain adaptation}

\begin{figure}[!t]
    \centering
    \includegraphics[width=\textwidth]{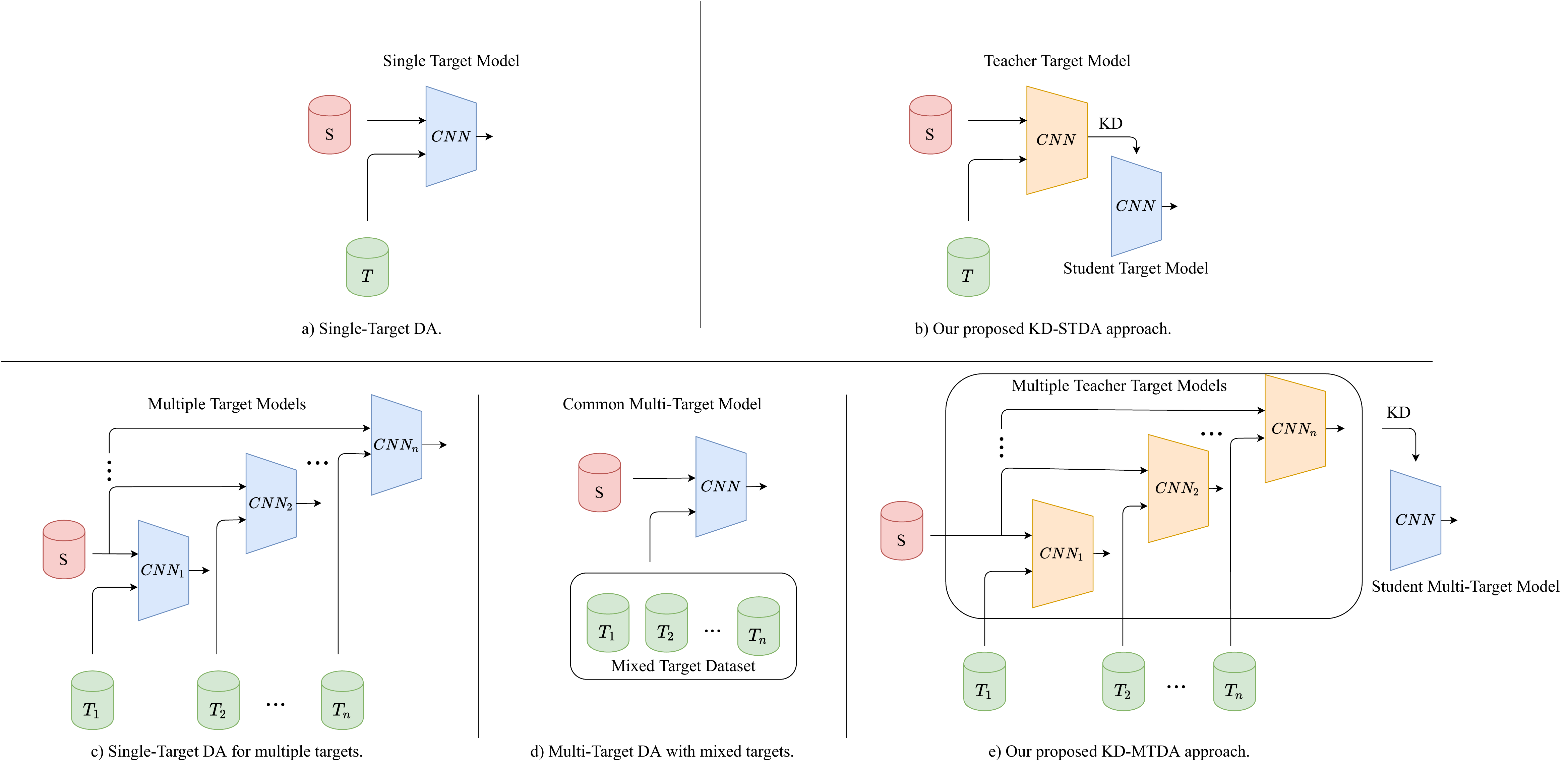}
    \caption{Illustration of the different strategies for STDA and MTDA of CNNs. a) Basic STDA technique given data from a source $S$ and target $T$ domain. b) Our proposed KD-STDA approach, where a teacher CNN is adapted to one target domain, and its knowledge is transferred to a smaller student CNN. c) MTDA by applying STDA independently on $n$ CNNs, one per target domain dataset. d) Applying STDA using a mixture of datasets from $n$ target domain to adapt a common CNN. e) Our proposed KD-MTDA approach, where multiple teacher CNNs are adapted, one per target, and their knowledge is combined and distilled to a smaller common student CNN.}
    \label{fig:DAvsMTDA}
\end{figure}

While single target domain adaptation (STDA) has gained significant attention, multi-target domain adaptation (MTDA) remains largely unexplored, since it poses a challenge for many applications. Recent work to tackle the MTDA task \cite{MTDA_Theoric} proposes to maximize the mutual information between domain labels and domain-specific features while minimizing the mutual information between shared features. Other papers have explored the open multi-domain adaptation \cite{compounddomainadaptation} or blended MTDA\cite{AMEAN, dada}. The first focuses on doing UDA on multiple target domains on a model so that it can also perform well on another unseen target domain not available during training. The second technique assumes that there's no information on where the target image comes from and uses a clustering technique to separate them into different pseudo target domains. The Figure \ref{fig:DAvsMTDA} shows the difference between our proposed technique and current existing techniques. 

\subsection{Joint unsupervised domain adaptation and knowledge distillation:}
Even though joint UDA and KD has already been investigated in the past \cite{KnowledgeAdaptation,SSDA_KD_MRI}, the aim of these works is on improving the UDA accuracy, neglecting the complexity reduction problem.
For example, \cite{KnowledgeAdaptation} uses multiple teachers to learn multiple source domains, increasing the accuracy of the student model for sentiment analysis. In \cite{SSDA_KD_MRI}, KD and DA are combined in a semi-supervised fashion in the context of medical image segmentation. 
To the best of our knowledge, only the method in \cite{TCP} currently tackles the problem of compression and domain shift using pruning and UDA. However, unlike the proposed model, the approach in \cite{TCP}, i.e., TCP, operates in several steps. TCP must first be adapted to a target domain, and then the learned model is iteratively pruned while continuously refined using UDA. This contrasts to our model, which progressively compresses and adapts to multiple target domains simultaneously from a single source CNN model.
Furthermore, TCP presents some important limitations. First, by performing the UDA and compression as two independent steps, there is a risk of a degrade in performance and a limited compression rate. Second, longer deployment times are needed for obtaining an efficient CNN that is adapted to a target domain, which can hamper the application to many specific problems, for example application that have a lot of changes in environment such as mobile car detection. Lastly, TCP\cite{TCP} cannot be easily generalized to an MTDA setting, and applying TCP directly on a mixture of multiple target domains can result in poor performance since it has been shown in literature \cite{AMEAN} that directly apply an STDA technique on a mixture of multiple targets will only have limited performance.


\section{Progressive Domain Adaptation Through Knowledge Distillation}
 
In this paper, we simultaneously address the domain shift and compression problems to obtain an efficient CNN model that is adapted to a target domain. Instead of obtaining a compressed CNN directly from a previously-adapted CNN, we argue that it may be beneficial to lean an efficient CNN by teaching it how to progressively adapt to target domains. This would mean that our efficient model is less constrained to optimize since it does not start from a local minimum. Based on this idea, we propose an approach that relies on a larger teacher model to performs UDA for a target domain, while progressively transferring the knowledge of how to perform UDA to a smaller student. Both student and teacher models begin from a pre-trained CNN, instead of a model previously adapted to a target domain. 

The main pipeline of our method is illustrated in Figure \ref{fig:KD-STDA}. The rest of this section provides additional details on UDA techniques used in this paper, KD for both supervised and unsupervised settings, our joint CNN optimization for UDA and KD, and generalization to multiple domains.

\subsection{Unsupervised domain adaptation:} \label{UDA_section}

Our distillation of domain adaptation is UDA-agnostic, and can be integrated using any UDA approach. We apply our technique on popular discrepancy- and adversarial-based UDA approaches from the literature. First, a Maximum Mean Discrepancy (MMD)-based UDA approach\cite{MMD_ICLR}, that seeks to minimize the distance between target and source domain distributions in the features space. The choice of this technique for domain adaptation allows for fair comparison of compression of our technique with Transfer Channel Pruning (TCP) \cite{TCP}. Second, we use a technique called RevGrad \cite{GRL}, which relies on adversarial training in order to find a common representation by encouraging domain confusion between target and source domain. The rest of this section provides additional details on these two approaches. 
 
\subsubsection{MMD approach:} 

Let us define the labeled source dataset as $S = \{x_s, y_s\}$ and the unlabeled target dataset as $T = \{x_t\}$. The UDA loss for teacher CNN based on MMD \cite{MMD_ICLR, TCP} is :
\begin{equation}\label{eq:teacher_mmd}
	\begin{aligned}
	\mathcal{L}_{MMD}(\phi, T, S) = || \frac{1}{N_s} \sum_{x_i \in S} \phi(x_i) - \frac{1}{N_t} \sum_{x_j \in T} \phi(x_j)||_\mathcal{H}^2
	\end{aligned}
\end{equation}
where dataset $S$ contains $N_s$ samples (and labels), $T$ contains $N_t$ samples, $\phi$ is the teacher function that maps an input to a feature map, and $\mathcal{H}$ is the Reproducing Kernel Hilbert Space (RKHS) with Gaussian kernel. As in \cite{MMD_ICLR, TCP}, we add a supervised loss on the source domain dataset, and the overall UDA loss for the teacher CNN is:
\begin{equation}\label{eq:teacher_da_mmd}
	\begin{aligned}
	\mathcal{L}_{TDA}(\Phi, T, S) = \mathcal{L}_{MMD} + \gamma \mathcal{L}_{CE}(\Phi(x_s, 1), y_s)
	\end{aligned}
\end{equation}
where $\mathcal{L}_{CE}$ the supervised cross-entropy loss of the teacher model on the source domain data, $\gamma$ a trade-off hyper-parameter that follows the same variations as \cite{TCP}, and $\Phi$ the output of the teacher network with a soft-max of temperature value set to 1 (i.e. the regular soft-max).

\subsubsection{RevGrad approach:}
 
For this approach, UDA of the teacher relies on a domain classifier, a gradient reversal layer (GRL), and the domain confusion loss defined as:
\begin{equation}\label{eq:teacher_dc}
\footnotesize
	\begin{aligned}
	\mathcal{L}_{DC}(\phi, T, S) = \frac{1}{N_s + N_t} \sum_{x \in S \cup T}\mathcal{L}_{CE}(C(\phi(x)), d_l)
	\end{aligned}
\end{equation}
where $\phi(x)$ is the output from the feature extractor of teacher network $\Phi$ (before the fully connected layers), $C$ is the domain classifier for the corresponding teacher network, $d_l$ the domain label (source or target), $N_s$ is the number of samples in the source domain $S$, and $N_{ti}$ is the number of samples in the target domain $T_i$. The overall UDA loss is then defined as:
\begin{equation}\label{eq:teacher_da_grl}
\footnotesize
	\begin{aligned}
	\mathcal{L}_{TDA}(\Phi, T, S) = \frac{1}{N_s}\sum_{x_s, y_s \in S}\mathcal{L}_{CE}(\Phi(x_s),y_s) + \alpha \cdot \mathcal{L}_{DC}(\phi, T, S)
	\end{aligned}
\end{equation}
The cross-entropy loss term in Equ.\ref{eq:teacher_da_grl} allows the supervised training of the teacher model on the source domain data to ensure the consistency of domain confusion. The second term is controlled by a hyper-parameter $\alpha$ that regulates the importance of the domain confusion loss which is maximized using a gradient reversal layer. 

 \subsection{Knowledge distillation for domain knowledge transfer:}
 
 The next step consists in transferring target domain knowledge from the teacher to student models. Our proposed method is general, and can be adapted to any KD method based on logits and features. We apply our proposed approach on a KD technique based on (1) the work of \cite{HIntonKD} that distill knowledge from a teacher to student using the output of each network, and (2) the feature distillation method \cite{Overhaul} that minimizes the difference between feature maps of intermediate layers. These KD techniques are detailed below.

\subsubsection{Logits-based distillation:} \label{Logits}
 
Let us defined the temperature based softmax that inputs the logits, and produce a softened output that can represent  more information to be distilled:
\begin{equation}\label{eq:softmax_temp}
	\begin{aligned}
	softmax_i(\tau) = \frac{\exp{(z_i \cdot \tau)}}{\sum_j\exp{(z_j \cdot \tau)}}
	\end{aligned}
\end{equation}
where $z_i$ the logits of class $i$ produced by the classification layer of the CNN, the Hinton distillation \cite{HIntonKD}, is then defined as:
\begin{equation}\label{eq:logits_distill}
	\begin{aligned}
	\mathcal{L}_{LDistill}(\Phi, \Theta, D) = \mathcal{L}_{KL}(\Theta(D, \tau), \Phi(D, \tau))
	\end{aligned}
\end{equation}
where $\Theta(D, \tau)$ and $\Phi(D, \tau)$ represent the outputs of the student and teacher networks, respectively, with a softmax based on a temperature $\tau$ in order to soften the output. For simplicity, $D$ represents a generic dataset that we will replace with the source or target dataset. 

\subsubsection{Feature-based distillation:}

Instead of distilling only on the output of teacher and student, feature KD \cite{Overhaul} relies on several intermediate layers inside the network. For simplicity, we present the KD loss for only one layer since the algorithm similar across all layers. For this technique, the features for distillation is extracted at the end of each residual block and before a ReLu unit since the information before ReLU still carries negative values which are important for KD. Additionally, the authors argue that not all negative values are important, therefore they propose applying a margin ReLU when distilling information from teacher to student. The margin ReLU is defined as:
\begin{equation}\label{eq:margin_relu}
	\begin{aligned}
	\sigma_m(x) = \max(x,m)
	\end{aligned}
\end{equation}
where $m$ is a margin value less than zero. The margin $m$ can be computed as the expectation value of the negative response, for each input channel. The margin value for channel \emph{c} can be computed as:
\begin{equation}\label{eq:margin_relu_channel}
	\begin{aligned}
	m_c = \emph{E}[F^{i}_{\Phi}|F^{i}_{\Phi} < 0, i \in C]
	\end{aligned}
\end{equation}
with $F^{i}_{\Phi}$ the teacher's feature of the $i-th$ channel of $C$ channel. The margin value \emph{$m_c$} is computed over all training samples. As for the distillation loss, the authors use a partial L2 distance function, defined as:
\begin{equation}\label{eq:feature_distance}
	\begin{aligned}
    \mathcal{L}_{PL2}(F_{\Phi}, F_{\Theta}) = \sum_{j}^{W H C}
    \begin{cases}
      0, & \text{if}\ F_{\Theta_j} \leq F_{\Phi_j} \leq 0 \\
      (F_{\Phi_j} - F_{\Theta_j})^2 , & \text{otherwise}
    \end{cases}
	\end{aligned}
\end{equation}

In this Equ. \ref{eq:feature_distance}, $F_{\Theta}, F_{\Phi} \in \mathcal{R}^{W \times H \times C}$ represents respectively the feature map of the student and teacher networks. Here, the $j-th$ component represents an element in the feature map where $F_{\Theta_j}, F_{\Phi_j} \in \mathcal{R}$. After the domain adaptation in \ref{UDA_section}, the target domain knowledge is transferred from the teacher to the student, we use a loss function:
\begin{equation}\label{eq:distill}
	\begin{aligned}
	\mathcal{L}_{FDistill}(\Phi, \Theta, D) = \mathcal{L}_{PL2}(\sigma_{m_c}(F_{\Phi}(D)), r(F_{\Theta}(D)
	\end{aligned}
\end{equation}
Here, $\sigma_{m_c}$ represents margin ReLU and r(·) represents 1×1 convolution layer. $F^T$ and $F^S$ represent respectively the features of student network and teacher network. In the next section, we will further explain our contribution that allows employing unsupervised KD for UDA.

\subsection{Progressive Joint KD and UDA: KD-STDA}

\begin{figure}[htbp]
    \centering
    \includegraphics[width=0.75\textwidth]{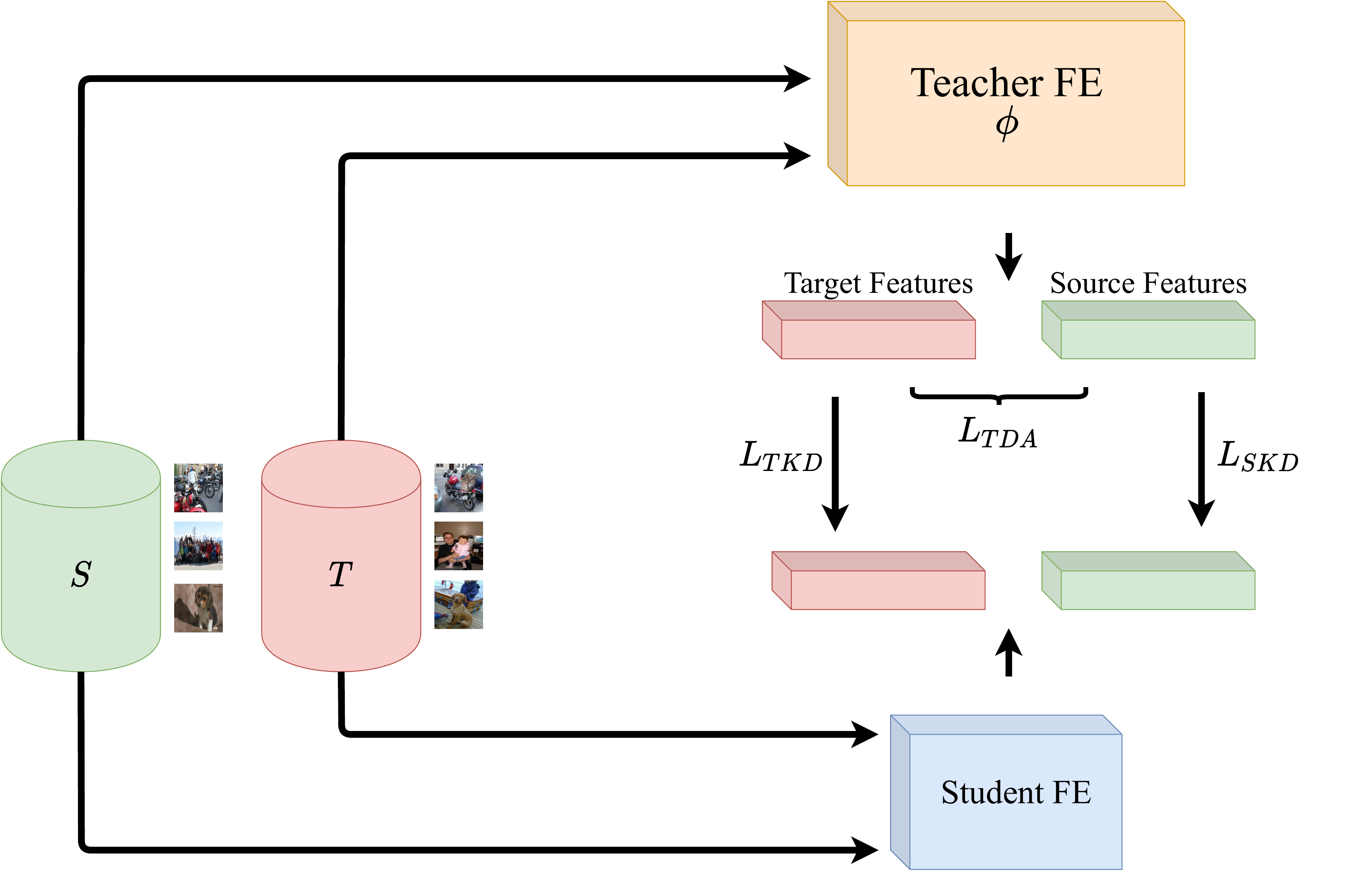}
    \caption{Our proposed progressive KD-STDA, where DA is performed on a teacher CNN, and then a smaller student CNN learns how to adapt on both target and source features through knowledge distillation.}
    \label{fig:KD-STDA}
\end{figure}

The main pipeline of the proposed method is depicted in Figure \ref{fig:KD-STDA}. Our method performs UDA of a teacher model by learning a common domain-invariant feature embedding between the source and target domains, while progressively distilling its knowledge to a student model. Contrary to other works, e.g., \cite{TCP}, the teachers in our setting are not adapted to the target domain, and start from pre-trained weights on ImageNet. Thus, our teachers would train the student CNN to adapt to the target domain, while they also continuously adapt to each target domain. From Equation $6$, target distillation is a simple application of the equation on the target dataset:
\begin{equation}\label{eq:target_distill}
	\begin{aligned}
	\mathcal{L}_{TKD}(\Phi, \Theta, T) = \mathcal{L}_{Distill}(\Phi, \Theta, T)
	\end{aligned}
\end{equation}

Equ.\ref{eq:target_distill} should allow to transfer information to the student network, since we are only interested in performing correctly on the target domain data. Inspired by current UDA techniques, we propose to add a consistency distillation loss in order to ensure that our student learns a common representation. This consistency is achieved by distilling source domain information into the student CNN:
\begin{equation}\label{eq:source_kd}
	\begin{aligned}
	\mathcal{L}_{SKD}(\Phi, \Theta, S) = \mathcal{L}_{distill}(\Phi, \Theta, S) + \alpha  \mathcal{L}_{CE}(S)
	\end{aligned}
\end{equation}

Eq. \ref{eq:source_kd} is the student KD loss, with hyper-parameter $\alpha$ to balance between the KD and the cross entropy loss of the output of the student model and the ground truth on the source domain. We also propose to add the $\beta$ hyper-parameter in order to balance out the importance between UDA and KD. Since we are performing jointly KD and DA, in the beginning, the teacher would still be learning from the DA. This means that there is not much to be learned for the student model, besides the source representation which can be learned from the KD loss. In light of this, we propose to start by giving more importance to UDA in the beginning and gradually transfer the importance to KD basing $\beta$ on an exponential growth function between $[b, f]$, with b the starting value of $\beta$ and $f$ the end value. In order to define as exponential growth, we need to calculate a growth rate based on $b$ and $f$:
\begin{equation}\label{eq:growth_rate}
	\begin{aligned}
	g = \frac{\log(f / b)}{N_e}
	\end{aligned}
\end{equation}
where $N_e$ is the number of epochs and $g$ the growth rate. Given the growth rate, $\beta$ at epoch $t$ can be found as:
\begin{equation}\label{eq:update_beta}
	\begin{aligned}
	\beta_t = b \cdot \exp\{g \cdot t\}
	\end{aligned}
\end{equation}

Figure \ref{fig:KD-STDA} illustrates all these losses and also the proposed techniques. The final loss of our models, is then:
\begin{equation}\label{eq:total_loss}
	\begin{aligned}
	\mathcal{L}(\Phi, \Theta, S) = (1 - \beta)\mathcal{L}_{TDA}(\Phi, T, S) + \beta(\mathcal{L}_{TKD}(\Phi, \Theta, T) + \mathcal{L}_{SKD}(\Phi, \Theta, S))
	\end{aligned}
\end{equation}
Algorithm \ref{KD-STDA-Algo} described the steps of our approach for the MTDA setting. 
\begin{algorithm}[htbp]
\SetAlgoLined
\caption{KD-STDA: Our method in STDA Setting.}
\label{KD-STDA-Algo}
\SetKwInOut{Input}{input}
\SetKwInOut{Output}{output}
\SetKwInOut{Parameter}{parameter}
\Input{A teacher model $M_T$, a student model $M_S$, a source dataset $D^{Sup}_s$, a target dataset $D^{U}_t$}
\Output{A target adapted student model}

\For {$epoch\gets{1}$ \KwTo {$N_e$}} {
    \For {$x_s$ in $S$ and $x_t$ in $T$} {
        Optimize the $(1 - \beta)\mathcal{L}_{DA}$ for the teacher model \\
        Optimize the $\beta(\mathcal{L}_{TKD} + \mathcal{L}_{SKD})$ for the student model\\
        Update $\beta$ following Eq.\ref{eq:update_beta}
    }
    Evaluate the model\\
}
\end{algorithm}

\subsection{Progressive Multi-target KD and UDA: KD-MTDA}

\begin{figure}[htbp]
    \centering
    \includegraphics[width=1.0\textwidth]{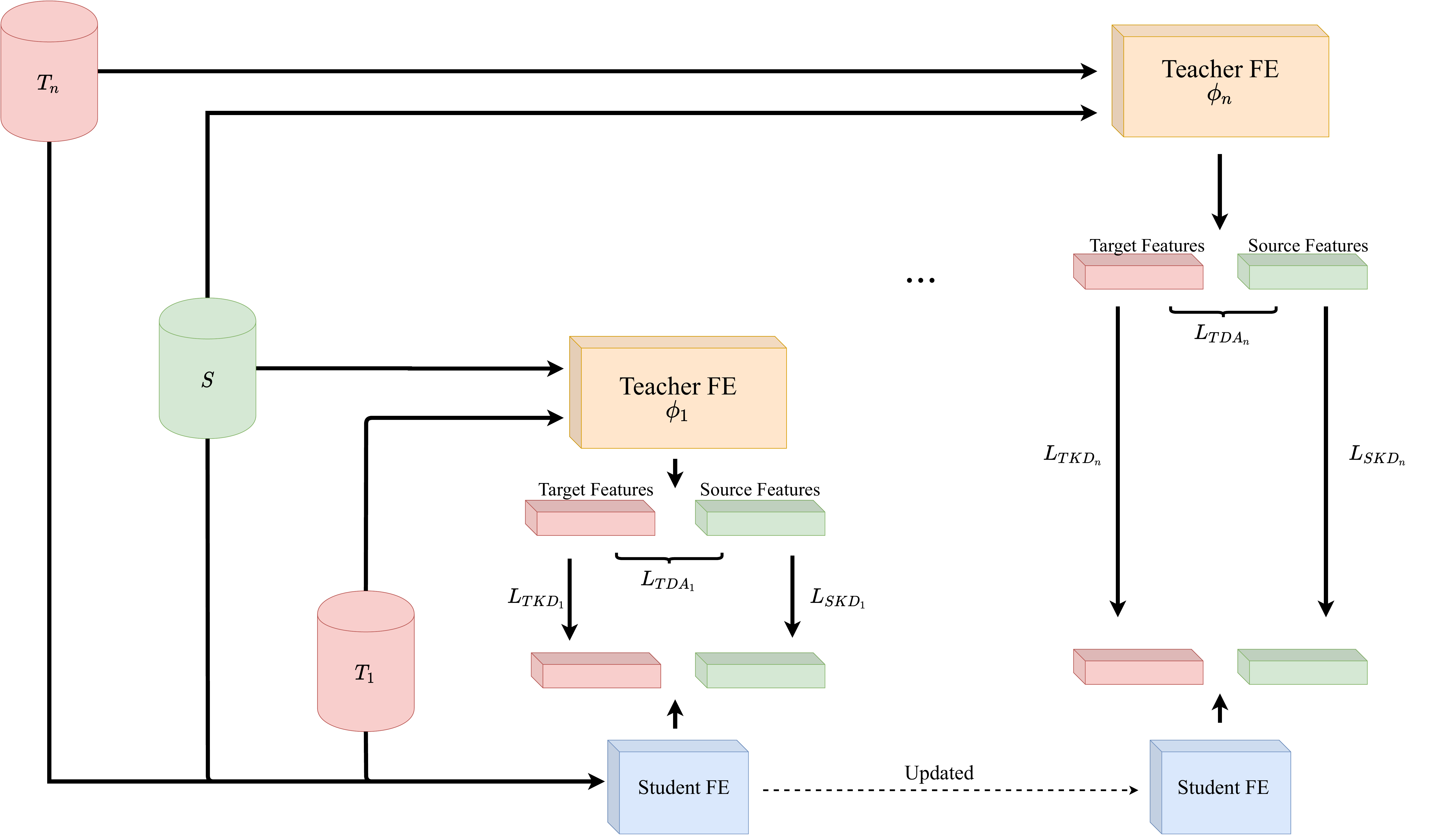}
    \caption{Our proposed method for MTDA with $n$ target domains. Each teacher CNN is assigned a different target domain, and sequentially teaches the student CNN how to adapt to source and target features for different target domains using knowledge distillation. After each knowledge distillation process with each teacher, features output from the student CNN are recalculated using update weights, for a new distillation with a different teacher.}
    \label{fig:MTDA}
\end{figure}

A multi-target approach can also be naturally derived from our approach by assigning each teacher a different target domain, and applying the same UDA loss that we previously defined. The progressive KD to the common student is described in Algorithm \ref{KD-STDA-MTDA-Algo}. This approach can be formalized in the same way as in the STDA setting. We define an ensemble of target datasets from $T_1$ to $T_{n}$, and redefine the loss to a specific target domain dataset for each teacher:
\begin{equation}\label{eq:Multi_target_loss}
	\begin{aligned}
	\mathcal{L}_{MTDA}(S, T_i) = (1 - \beta)\mathcal{L}_{TDA}(\Phi_i, T_i, S) + \beta(\mathcal{L}_{TKD}(\Phi_i, \Theta, T_{i}) + \mathcal{L}_{SKD}(\Phi_i, \Theta, S)
	\end{aligned}
\end{equation}
where $\Phi_1$ to $\Phi_n$ are the respective teachers for target domains $T_1$ to $T_{n}$. Similar to our STDA setting, we rely on the exponentially growing rate to gradually transfer the importance of domain adaptation to distillation since the student model and teachers are quite similar in the beginning (they all start from pre-trained weights of ImageNet). 

Figure \ref{fig:MTDA} illustrated the block diagram of our proposed KD-MTDA method. By assigning each complex teacher a target domain, we can adapt a CNN to generalize well across multiple target domains, thus allowing the student to learn progressively transferable features from all the teachers. 
\begin{algorithm}[htbp]
\footnotesize
\SetAlgoLined
\caption{KD-MTDA: Our method in MTDA setting.}
\label{KD-STDA-MTDA-Algo}
\SetKwInOut{Input}{input}
\SetKwInOut{Output}{output}
\SetKwInOut{Parameter}{parameter}
\Input{A source domain dataset $S$, a set of target dataset $T_0, T_1, ... T_n$}
\Output{A student model adapted to $n$ targets}

Initialize a set of teachers models $\Phi = \{\Phi_0, \Phi_1, ...\Phi_n\}$\\
Initialize a student model $\Theta$ \\

\For {$e\gets{1}$ \KwTo {$N_{e}$}} {
    \For {$x_s \in S$ and $X_t \in \{T_0,...T_n\}$} {
        Get the set of data of target domains $X_t$ \\
        \For {$x^{i}_{t} \in X_t$ and $\Phi_i \in \Phi$} {
            Optimize  $(1-\beta)\mathcal{L}_{DA}$ for $\Phi_i$ using $x_s, x_t^i$ \\
            Optimize the loss of equation $\beta\mathcal{L}_{SKD}$ for $\Phi_i$ and $\Theta$ using $x_s$ \\
            Optimize the loss of equation $\beta\mathcal{L}_{TKD}$ for $\Phi_i$ and $\Theta$ using $x_t^i$ \\
        }
        Update $\beta = s \cdot \exp^{g \cdot e}$
    }
    Evaluate the model\\
}
\end{algorithm}

\section{Experimental Methodology}


\subsection{Datasets:}

\paragraph{Digits} This dataset is made of a set of digits datasets: MNIST\cite{mnist} (\textbf{mt}), MNIST-M\cite{mnist-m} (\textbf{mm}), SVHN\cite{svhn} (\textbf{sv}), and USPS\cite{usps} (\textbf{up}). Each one has each 10 classes that represent all the digits. For the evaluation on this dataset, we follow the same protocol as in \cite{MTDA_Theoric} for a fair comparison: where we use 25000 samples for training on \textbf{mt},\textbf{mm},\textbf{sv},\textbf{sy} and 9000 for testing. On the \textbf{up} dataset we use the entire dataset as a domain. We will mainly use this dataset for comparison of MTDA. 

\paragraph{Office31}\cite{oFfice31} This dataset contains three subsets of dataset which are Webcam (\textbf{W}), DSLR (\textbf{D}) and Amazon (\textbf{A}). These subsets contain images from amazon.com (A) or office product taken a DSLR camera (D) or a webcam (W). We evaluate our results based on the same scenario as \cite{TCP}. These datasets all have 31 common classes and around 4000 images in total and each subset has a different amount of data: Amazon (2800 images), DSLR (500 images), Webcam (800 images).

\paragraph{ImageClef-DA} This dataset for UDA has four subsets which are taken from Imagenet (\textbf{I}), Pascal-Voc (\textbf{P}), Caltech (\textbf{C}) and Bing (\textbf{B}). Each of these subsets contains 600 images for 12 classes. For this dataset, similars to Office31, we use the same scenario as \cite{TCP} for fair comparison.

\paragraph{PACS}\cite{pacs} While this dataset is often used for domain generalization, it has been used in \cite{MTDA_Theoric} for MTDA. This dataset contains 4 subsets -- Art painting (\textbf{Ap}), Photo (\textbf{P}), Cartoon (\textbf{Cr}) and Sketch (\textbf{S}). With around \~2000 images in Art Painting, \~2300 for Cartoon, \~1700 for Photo and \~4000 for Sketch. We will use this dataset for MTDA comparison. 

\paragraph{OfficeCaltech} This dataset contains 4 subsets of four different domains: Amazon (\textbf{A}), Caltech10 (\textbf{C}), DSLR(\textbf{D}) and Webcam (\textbf{W}). They each contain respectively 957, 1123, 295, 157 images of 10 classes.

\begin{figure}[htbp]
    \centering
    \includegraphics[width=\textwidth]{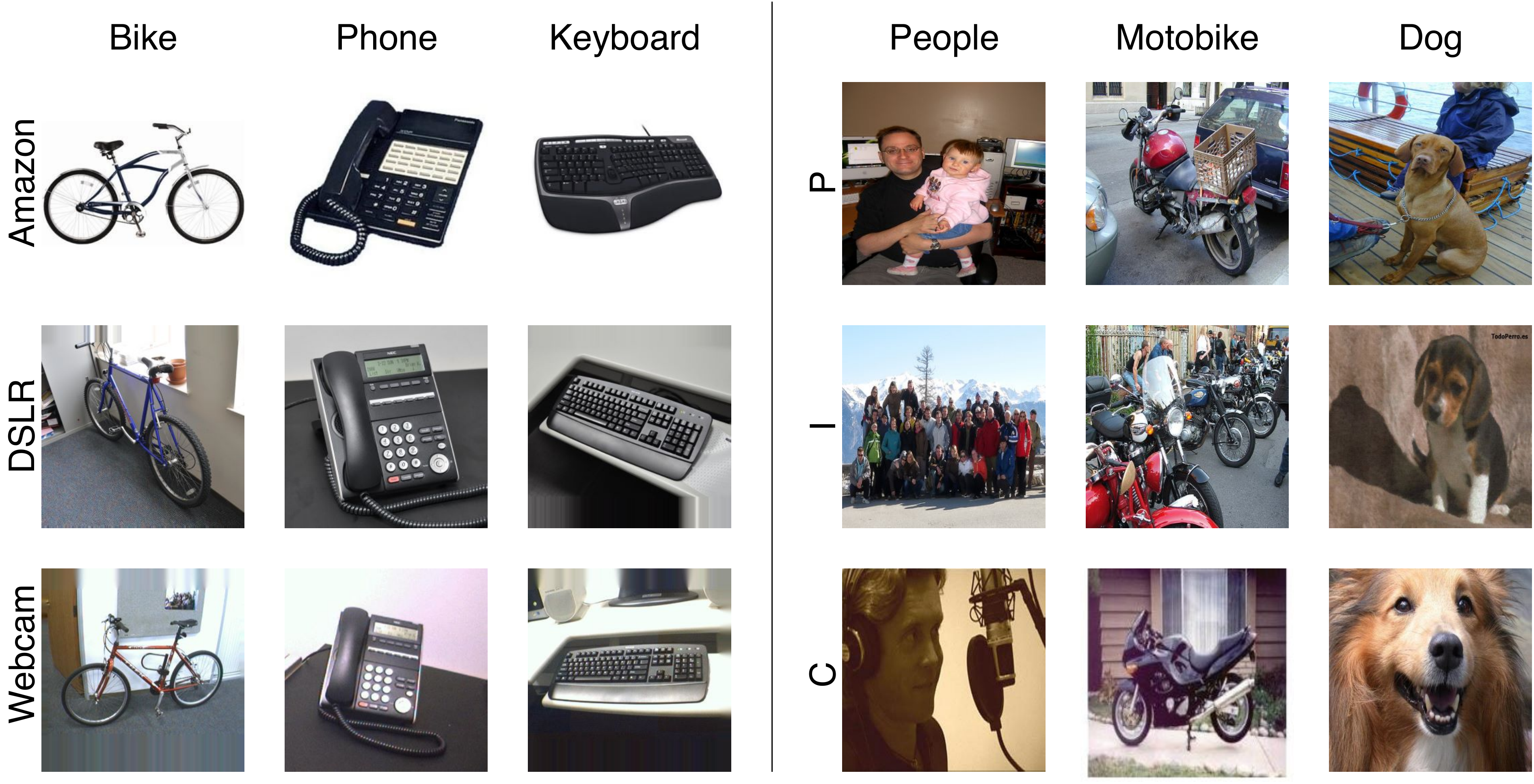}
    \caption{Samples from the Office31 and ImageClef-DA datasets.}
    \label{fig:datasets}
\end{figure}

\subsection{Baselines methods:}


For experiments in the STDA setting, several baselines were selected in order to evaluate our method. We start with the three scenarios presented earlier: (1) UDA then KD, (2) KD then UDA and (3) UDA directly on compact model. For these baselines, our teacher model will always be ResNet50 and our student models ResNet18 or ResNet34. In the first case, $UDA \xrightarrow{} KD$, UDA is applied to a teacher model then we start KD to a student model on target domain without labels. For the second baseline $KD \xrightarrow{} UDA$, the teacher model is first trained in supervised way on the source dataset then we apply KD with this teacher and a student using labeled data of the source dataset. The student is then applied UDA. As for the last baseline, UDA is applied directly to a student model. We also add RevGrad\cite{GRL} and CDAN\cite{CDAN} directly on compact model as another baseline. In addition, we also compare to the state-of-the-art method in compression and UDA: TCP\cite{TCP}. 

In the MTDA setting, we compare our method with several baselines like a lower-bound where the model is only trained on source domain and then test on multiple target domains. We also introduce baselines from current STDA techniques, RevGrad\cite{GRL} and ADDA\cite{ADDA} to show that, in their current form, they are not suitable for an MTDA setting. Additionally, we also compare our method with \cite{MTDA_Theoric} which is the current state-of-the-art MTDA technique that uses domain labels. As for \cite{compounddomainadaptation}, we do not include them in our comparison since it focuses on the open MTDA with unseen target domain, however we provide a comparison with DADA\cite{dada} which does use domain labels.

\subsection{Implementation details:}

For the STDA setting, we implement our method with two optimizers for domain adaptation and knowledge distillation. For our experiments on Office31 and ImageClef-DA, in order to have a fair comparison to TCP, our teacher model was chosen such that it's the same starting model as TCP and our student models have similar number of FLOPS or less than TCP pruned models. These FLOPS are measured for our models, as for the FLOPS of TCP, we report the number used in their paper. In addition, in order to show that's our framework can work with most domain adaptation and distillation techniques, we provide the result using combination of logits and feature distillation with MMD and RevGrad domain adaptation.

For the implementations of MTDA setting, we use the same number of optimizers as teacher models, with each optimizer responsible for the domain adaptation. Similar to our STDA setting, we have another optimizer for the knowledge distillation of the student. For this approach, we mainly use RevGrad and logits as domain adaptation and distillation technique. In order to have a fair comparison with state-of-the-art method \cite{MTDA_Theoric}, we decided to use LeNet as student backbone and ResNet50 as teachers. Since the authors use a non-standard backbone, with multiple residual layers and deconv layers, we decided to use a standard backbone (LeNet) that's close to their backbone while having less computational power for fair comparison. For the accuracy, we report the average accuracy of our model across all target domains. 

For both KD-STDA and KD-MTDA, the details for our hyper-parameters can be found in the supplementary material. Both methods were implemented in PyTorch, and experiments were run on a server equipped with a Nvidia P100 GPU.


\section{Results and Discussions}


\subsection{STDA with logits- and feature-based distillation:}

Table \ref{tb:Office31_stda_result_34} shows the accuracy of UDA methods on the Office31 dataset. Results indicate that our KD-STDA method outperforms all the baselines and state-of-the-art TCP \cite{TCP}. As for the comparison with our baseline scenarios, the second baseline $KD \xrightarrow{} UDA$ has the best performance among all the scenarios and better than the state-of-the-art. The difference between the second and third baselines "$UDA$ only" is because the compact model of the second baseline has distilled features which proves to be the better start than the third baseline. The first baseline $UDA \xrightarrow{} KD$ provides the lowest accuracy due to unsupervised knowledge distillation on the target domain without a consistency loss. In addition, our technique on both student models performs better than state-of-the-art TCP, especially on ResNet18, where our method improved the most upon state-of-the-art. Finally, our progressive method also seems to improve upon UDA since we perform better than the second baseline.

\begin{table*}[b!]
\caption{Target domain accuracy of the proposed and baseline UDA methods on the Office31 dataset, when ResNet50 is the teacher, and Resnet34 and ResNet18 are the desired (student) CNN backbones. Results are shown for logits- and feature-based distillation: LB $\vert$ FB.}
\label{tb:Office31_stda_result_34}
\centering
\resizebox{\textwidth}{!}{
\begin{tabular}{|l||r|r|r|r|r|r||r|}
\hline
    & \multicolumn{7}{|c|}{\textbf{Source $\rightarrow$ Target}}  \\ 
\textbf{Training methods}           & \textbf{A $\xrightarrow{}$ W} & \textbf{W $\xrightarrow{}$ A} & \textbf{D $\xrightarrow{}$ W} & \textbf{W $\xrightarrow{}$ D} & \textbf{D $\xrightarrow{}$ A} & \textbf{A $\xrightarrow{}$ D} & \textbf{Average}     \\ \hline \hline
\multicolumn{8}{|c|}{\textbf{Teacher: ResNet50 --- Student: ResNet34}} \\ \hline
Baseline 1: UDA $\xrightarrow{}$ KD & 25.4  $\vert$ 9.6   & 7.1 $\vert$ 6.3     &  28.5 $\vert$ 12.6   & 50.0  $\vert$  13.8   & 9.7 $\vert$ 5.3    & 30.7 $\vert$ 11.2   & 25.2 $\vert$ 9.8        \\ \hline
Baseline 2: KD $\xrightarrow{}$ UDA & 75.7 $\vert$ 51.8 & 61.2 $\vert$ 22.0 & 97.8 $\vert$ 78.3 & 99.7 $\vert$ 93.2 & 59.6 $\vert$ 17.3 & 81.1 $\vert$ 57.2 & 79.1 $\vert$ 53.3 \\ \hline
Baseline 3: UDA only on ResNet34        & 67.2   & 52.3   & 93.6   & 96.6   & 52.2   & 71.6   & 72.2       \\ \hline \hline
RevGrad & 75.2 & 59 & 97.4 & 99.8 & 59.1 & 78.0 & 78.1 \\ \hline
CDAN & 76.2 & 59.2 & 98.0 & 99.6 & 59.4 & 78.9 & 78.5 \\ \hline
TCP: prune rate = 12\% & 81.8   & 55.5   & 98.2   & 99.8   & 50.0     & 77.9   & 77.2        \\ \hline
KD-STDA MMD (Ours)  & 85.7 $\vert$ \textbf{86.0}   & 62.3 $\vert$ \textbf{67.6}   & 97.1 $\vert$ \textbf{99.0}   & \textbf{100} $\vert$ \textbf{100}    & 61.8 $\vert$ \textbf{66.4}   & 82.1 $\vert$ \textbf{84.7}   & 81.5 $\vert$ \textbf{83.9} \\ \hline \hline
%
\multicolumn{8}{|c|}{\textbf{Teacher: ResNet50 --- Student: ResNet18}} \\ \hline
Baseline 1: UDA $\xrightarrow{}$ KD  & 28.8 $\vert$ 10.2   & 5.8 $\vert$ 5.3    & 33.7 $\vert$ 11.5   & 51.1 $\vert$ 12.7   & 7.8 $\vert$ 7.8    & 25.1 $\vert$ 9.2   & 25.3 $\vert$ 9.4        \\ \hline
Baseline 2: KD $\xrightarrow{}$ UDA & 69.0 $\vert$ 53.1 & 57.3 $\vert$ 16.8 & 96.2 $\vert$ 82.3 & \textbf{100} $\vert$ 88.6 & 56.3 $\vert$ 19.8 & 73.6 $\vert$ 54.4 & 75.4 $\vert$ 52.5 \\ \hline
Baseline 3: UDA only on ResNet18        & 60.2   & 49.2   & 93.7   & 97.7   & 47.6   & 66.4   & 69.1       \\ \hline \hline
RevGrad & 71.6 & 53.4 & 96.5 & 99.2 & 53.6 & 75.5 & 75.0 \\ \hline
CDAN & 72.3 & 53.9 & 96.9 & 99.3 & 53.9 & 76.1 & 75.4 \\ \hline
TCP: prune rate = 45\% & 77.4   & 46.3   & 96.3   & \textbf{100}   & 36.1     & 72.0   & 71.3         \\ \hline
KD-STDA MMD (Ours)        & 78.9 $\vert$ \textbf{87.3}   & 58.1 $\vert$ \textbf{64.1}   & 94.2 $\vert$ \textbf{99.0}   & \textbf{100} $\vert$ \textbf{100}    & 57.2 $\vert$ \textbf{63.3}   & 81.7 $\vert$ \textbf{85.7}   & 77.8 $\vert$ \textbf{83.2} \\ \hline
\end{tabular}
}
\end{table*}

Table \ref{tb:ImageClef_stda_result_34} shows the accuracy of methods on ImageClef, confirming the tendencies of Table \ref{tb:Office31_stda_result_34}. Our third baseline "$UDA$ only" performs better than TCP in this dataset, which can explain why our methods performance better. Both Tables \ref{tb:Office31_stda_result_34} and \ref{tb:ImageClef_stda_result_34} show that our techniques perform well with our proposed feature-based KD method. Results follow our previous analysis concerning the different baselines. In addition, feature-based distillation can perform better than logits-based distillation, as was shown in \cite{Overhaul}.

\begin{table*}[t!]
\caption{Same results as in Tables \ref{tb:Office31_stda_result_34} but on the ImageClef-DA dataset.}
\label{tb:ImageClef_stda_result_34}
\centering
\resizebox{\textwidth}{!}{
\begin{tabular}{|l||r|r|r|r|r|r||r|}
\hline
& \multicolumn{7}{|c|}{\textbf{Source $\rightarrow$ Target}}  \\ 
\textbf{Training methods}            & \textbf{I $\xrightarrow{}$ P} & \textbf{P $\xrightarrow{}$ I} & \textbf{I $\xrightarrow{}$ C} & \textbf{C $\xrightarrow{}$ I} & \textbf{C $\xrightarrow{}$ P} & \textbf{P $\xrightarrow{}$ C} & \textbf{Average} \\ \hline \hline
\multicolumn{8}{|c|}{\textbf{Teacher: ResNet50 --- Student: ResNet34}} \\ \hline
Baseline 1: UDA $\xrightarrow{}$ KD            & 48.8 $\vert$ 12.2       & 41.0 $\vert$ 14.3       & 46.0 $\vert$ 12.2       & 39.6 $\vert$ 11.5       & 38.8 $\vert$ 9.4       &    39.0 $\vert$ 8.7    & 42.0 $\vert$ 11.4        \\ \hline
Baseline 2: KD $\xrightarrow{}$ UDA & 76.6 $\vert$ 70.2 & 87.3 $\vert$ 77.3 & 92.0 $\vert$ 88.2 & 80.0 $\vert$ 73.2 & 65.6 $\vert$ 60.6 & 90.3 $\vert$ 84.2 & 81.9 $\vert$ 75.6 \\ \hline
Baseline 3: UDA only on ResNet34   & 73.3  & 86.3   & \textbf{92.6}   & 79.3   & 65.8   & 87.5   &    80.8     \\ \hline \hline
RevGrad & 75.9 & 87.0 & 92.2 & 79.3 & 65.3 & 90.0 & 81.6 \\ \hline
CDAN & \textbf{76.8} & 87.9 & 93.3 & 80.2 & 65.9 & 90.2 & 82.4 \\ \hline
TCP: prune rate = 12\% & 75.0  & 82.6   & 92.5   & 80.8   & 66.2   & 86.5   & 80.6    \\ \hline
KD-STDA MMD (Ours) & 75.6 $\vert$ 72.2  & \textbf{89.0} $\vert$ 87.5   & \textbf{92.6} $\vert$ 92.2   & \textbf{83.8} $\vert$ 82.5   & 66.5 $\vert$ \textbf{66.8}   & \textbf{92.8} $\vert$ 89.5   & \textbf{83.3} $\vert$ 81.8    \\ \hline \hline
\multicolumn{8}{|c|}{\textbf{Teacher: ResNet50 --- Student: ResNet18}} \\ \hline
Baseline 1: UDA $\xrightarrow{}$ KD             & 45.1 $\vert$ 10.3       & 41.8 $\vert$ 12.2       & 42.5 $\vert$ 14.2       & 43.1 $\vert$ 15.5       & 43.3 $\vert$ 7.7       &    34.5 $\vert$ 6.8    & 41.7 $\vert$ 11.1             \\ \hline
Baseline 2: KD $\xrightarrow{}$ UDA & 72.1 $\vert$ 68.3 & 86.3 $\vert$ 72.3 & 91.8 $\vert$ 84.6 & 74.6 $\vert$ 71.9 & 61.8 $\vert$ 62.1 & 90.6 $\vert$ 85.1 & 79.5 $\vert$ 74.0 \\ \hline
Baseline 3: UDA only on ResNet18   & 70.6   & 83.8   & 86.1   & 75.3   & 62.0     & 89.1   & 77.8 \\ \hline \hline
RevGrad & 71.2 & 86.8 & 92.0 & 76.7 & 63.9 & 89.9 & 80.1 \\ \hline
CDAN & 72.1 & 87.1 & \textbf{92.3} & 77.8 & 63.7 & 89.5 & 80.4 \\ \hline
TCP: prune rate = 45\%  & 67.8   & 77.5   & 88.6   & 71.6   & 57.7   & 79.5   & 73.7 \\ \hline
KD-STDA MMD (Ours) & 73.1 $\vert$ \textbf{73.8}   & 88.0 $\vert$ \textbf{88.2}   & 92.1 $\vert$ 91.5   & 77.3 $\vert$ \textbf{78.7}   & \textbf{65.6} $\vert$ 64.4   & \textbf{91.0} $\vert$ 89.9   & \textbf{81.1} $\vert$ \textbf{81.1}     \\ \hline
\end{tabular}
}
\end{table*}

\begin{figure}[b!]
    \centering
    \includegraphics[width=0.8\textwidth]{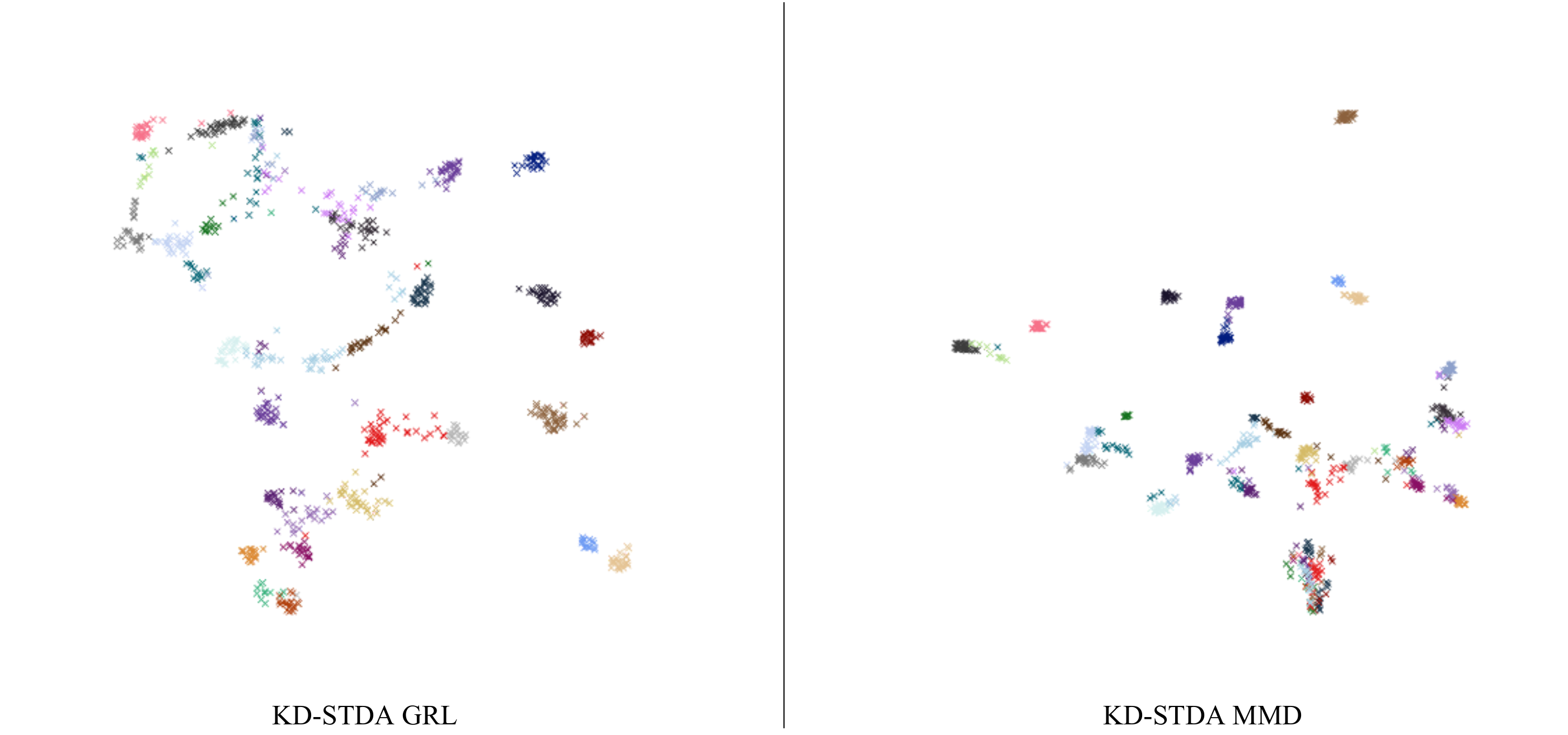}
    \caption{t-SNE visualisation of RevGrad and MMD with KD-STDA for scenario A $\xrightarrow{}$ W on ResNet34. Best viewed in color. We provide higher resolution figures in Supplementary Material.}
    \label{fig:tsne_revgrad_mmd}
\end{figure}

Up until now, we have employed an discrepancy-based (MMD) method for STDA. For this experiment, we replace this method with an adversarial-based method. We choose the popular adversarial method with GRL called RevGrad \cite{GRL} as  since it has been used and adapted to other tasks like object detection, image segmentation. Table \ref{tb:Office31_result_34_mmd_grl} shows the difference in accuracy between using RevGrad\cite{GRL} and MMD\cite{MMD_ICLR}. Results indicate that RevGrad provides similar accuracy compared to MMD based method with both logits and feature distillation. In Figure \ref{fig:tsne_revgrad_mmd}, the t-SNE visualization \cite{tsne}, computed from the output of the feature extractor of each CNN, of the scenario A $\xrightarrow{}$ W on ResNet34, further confirms the similarity between both variants.

\begin{table*}[ht]
\caption{Same results as in Tables \ref{tb:Office31_stda_result_34} but using MMD and GRL methods.}
\label{tb:Office31_result_34_mmd_grl}
\centering
\resizebox{\textwidth}{!}{
\begin{tabular}{|l||r|r|r|r|r|r||r|}
\hline
& \multicolumn{7}{|c|}{\textbf{Source $\rightarrow$ Target}}  \\ 
\textbf{Training methods}           & \textbf{A $\xrightarrow{}$ W} & \textbf{W $\xrightarrow{}$ A} & \textbf{D $\xrightarrow{}$ W} & \textbf{W $\xrightarrow{}$ D} & \textbf{D $\xrightarrow{}$ A} & \textbf{A $\xrightarrow{}$ D} & \textbf{Average}     \\ \hline \hline
\multicolumn{8}{|c|}{\textbf{Teacher: ResNet50 --- Student: ResNet34}} \\ \hline
KD-STDA MMD & 85.7 $\vert$ \textbf{86.0}   & 62.3 $\vert$ \textbf{67.6}   & 97.1 $\vert$ \textbf{99.0}   & \textbf{100} $\vert$ \textbf{100}    & 61.8 $\vert$ \textbf{66.4}   & 82.1 $\vert$ \textbf{84.7}   & 81.5 $\vert$ \textbf{83.9} \\ \hline
KD-STDA RevGrad & \textbf{82.6} $\vert$ 82.1   & \textbf{64.4} $\vert$ 63.1   & 97.8 $\vert$ \textbf{98.7}   & \textbf{100} $\vert$ \textbf{100}    & 64.0 $\vert$ \textbf{64.2}   & \textbf{86.0} $\vert$ 85.3   & \textbf{82.5} $\vert$ 82.2 \\ \hline

\end{tabular}
}
\end{table*}

\subsection{MTDA with logits-based distillation:}

\begin{table}[htbp]
\centering
\caption{Target accuracy of the proposed and reference MTDA methods on the Digits dataset.}
\label{tb:digits4_comp_mtda}
\resizebox{\textwidth}{!}{
\begin{tabular}{|l|r|r|r|r|r|r|r|r|}
\hline
& \multicolumn{8}{|c|}{\textbf{Source $\rightarrow$ Targets}}  \\ 
\textbf{LeNet}      & \multicolumn{4}{c|}{\textbf{sv $\rightarrow$ mt, mm, up}}                         & \multicolumn{4}{c|}{\textbf{mm $\rightarrow$ sv, mt, up}}                          \\ 
            & \textbf{sv $\rightarrow$ mt} & \textbf{sv $\rightarrow$ mm} & \textbf{sv $\rightarrow$ up} & \textbf{Average} & \textbf{mm $\rightarrow$ sv} & \textbf{mm $\rightarrow$ mt} & \textbf{mm $\rightarrow$ up} & \textbf{Average} \\ \hline \hline
\multicolumn{9}{|c|}{\textbf{Teacher: ResNet50 --- Student: LeNet}} \\ \hline
Source Only & 62.1               & 40.4               & 39.9              & 47.5    & 40.0               & 84.5               & 80.4               & 68.3    \\ \hline
ADDA        & 77.7               & 64.2               & 64.1              & 68.7    & 40.6               & 92.8               & 80.7               & 71.4    \\ \hline
RevGrad     & 73.8               & 61.0               & 62.5              & 65.8    & 51.8               & 61.0               & 85.3               & 66.0    \\ \hline
MTDA-ITA\cite{MTDA_Theoric}    & 84.6               & \textbf{65.3}               & 70                & 73.3    & 53.5               & \textbf{98.2}               & \textbf{94.1}               & 81.9    \\ \hline \hline
KD-MTDA   & \textbf{85.1}               & 63.9               & \textbf{81.5}              & \textbf{76.8}    & \textbf{61.8}               & 97.8               & 86.4               & \textbf{82.0}    \\ \hline
\end{tabular}
}
\end{table}

Tables \ref{tb:digits4_comp_mtda} shows the accuracy of our generalized algorithm KD-MTDA for a multi-target setting on each different target domains and their average on Digits dataset. Results indicate that our method improves by $3\%$ upon the current state-of-the-art \cite{MTDA_Theoric} on $sv \rightarrow mt, mm, up$. As for the other scenario, we only perform slightly better, this is because we use the same hyper-parameter on different scenarios and on different teachers. One way of improving the current result would be to have a different set of hyper-parameters for each teacher such that it is best optimized for each combination of source and target. In addition, compared to a domain adaptation on a mixed target domains, MTDA focused algorithm perform much better overall.

\begin{table}[t!]
\centering
\caption{Same results as in Tables \ref{tb:digits4_comp_mtda} but on PACS dataset.}
\label{tb:pacs_comp_mtda}
\resizebox{0.8\textwidth}{!}{
\begin{tabular}{|l|l|l|l|l|l|l|l|l|}
\hline
& \multicolumn{8}{|c|}{\textbf{Source $\rightarrow$ Targets}}  \\ 
\textbf{LeNet}    & \multicolumn{4}{c|}{\textbf{P $\xrightarrow{}$ Ap, Cr, S}}                       & \multicolumn{4}{c|}{\textbf{Ap $\xrightarrow{}$ Cr, S, P}}                       \\ 
         & \textbf{P $\xrightarrow{}$ Ap} & \textbf{P $\xrightarrow{}$ Cr} & \textbf{P $\xrightarrow{}$ S} & \textbf{Average}  & \textbf{Ap $\xrightarrow{}$ Cr} & \textbf{Ap $\xrightarrow{}$ S} & \textbf{Ap $\xrightarrow{}$ P} & \textbf{Average}  \\ \hline \hline
\multicolumn{9}{|c|}{\textbf{Teacher: ResNet50 --- Student: LeNet}} \\ \hline
ADDA     & 24.3              & 20.1              & 22.4              & 22.3 & 17.8              & 18.9              & 32.8              & 23.2 \\ \hline
MTDA-ITA \cite{MTDA_Theoric} & \textbf{31.4}              & 23.0              & 28.2              & 27.6 & 27.0              & 28.9              & \textbf{35.7}              & 30.5 \\ \hline \hline
KD-MTDA  & 24.6              & \textbf{32.2}              & \textbf{33.8}              & \textbf{30.2} & \textbf{46.6}              & \textbf{57.5}              & \textbf{35.6}              & \textbf{46.6} \\ \hline
\end{tabular}
}
\end{table}
From Table \ref{tb:pacs_comp_mtda}, where we show our accuracy on the PACS dataset, in contrast to the previous comparison, our method perform significantly better than the current state-of-the-art \cite{MTDA_Theoric}. The results from this dataset also confirm our hypothesis that by applying current STDA technique on multiple target domains, it does not yield good result and the model fail to generalize across multiple target domains. In addition, the paper of \cite{MTDA_Theoric} use an architecture that includes several Residual block layers whereas, our architecture is based on LeNet5 which only consisted of 5 convolution layers, this meant that our method not only performs better than current method, it can also work with smaller models.

\begin{table*}[htbp]
\large
\centering
\caption{Accuracy of KD-MTDA and DADA\cite{dada} by using Alexnet (student CNN) and Resnet50 (teacher CNNs) as backbones on the Office-Caltech dataset.}
\label{tb:Officecaltech_comp}
\resizebox{0.8\textwidth}{!}{
\begin{tabular}{|l|r|r|r|r|r|}
	\hline
	\textbf{Models}          & \textbf{A $\xrightarrow{}$ C,D,W}  & \textbf{C $\xrightarrow{}$ A,D,W} & \textbf{D $\xrightarrow{}$ A,C,W} & \textbf{W $\xrightarrow{}$ A,C,D} & \textbf{Average} \\ \hline \hline
		\multicolumn{6}{|c|}{\textbf{Teacher: ResNet50 --- Student: AlexNet}} \\ \hline
	Source only             & 83.1        & 88.9      & 86.7      & 82.2          & 85.2         \\ \hline
	RevGrad\cite{GRL}       & 85.9        & 90.5      & 88.6      & 90.4          & 88.9         \\ \hline
	DADA\cite{dada}         & 86.3        & 91.7      & 89.9      & 91.3          & 89.8         \\ \hline \hline
	KD-MTDA           & \textbf{93.3}        & \textbf{93.9}      & \textbf{90.1}      & \textbf{91.2}          & \textbf{92.1}         \\ \hline 
\end{tabular}
}
\end{table*}

Results in  Table\ref{tb:Officecaltech_comp} show the accuracy CNNs trained with KD-MTDA on OfficeCaltech dataset compared with a MTDA technique that does not rely on domain labels. The results in this Table further validate the importance of our method.

\begin{figure}[t!]
    \centering
    \includegraphics[width=0.8\textwidth]{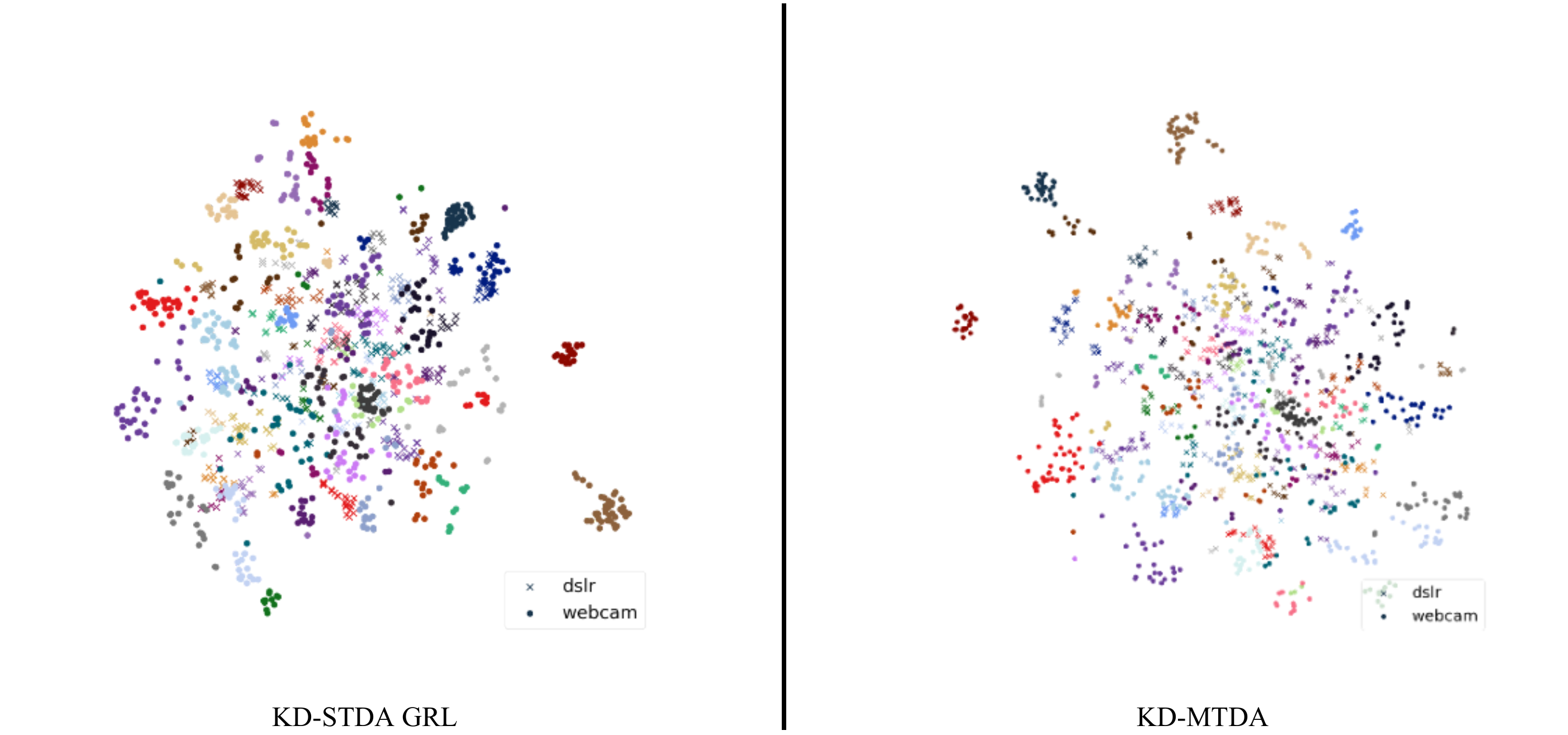}
    \caption{t-SNE of KD-STDA with mixed targets and KD-MTDA. Best viewed in color. We provide higher resolution figures in Supplementary Material.}
    \label{fig:tsne_stda_vs_mtda}
\end{figure}

Our method is also compared for a single vs. multiple teachers using same hyper-parameters as with the STDA with AlexNet as student CNN backbone.  From Table \ref{tb:Office31_stda_vs_mtda},  multiple teachers always outperform the single teacher whether using the domain labels or not. In addition, Figure \ref{fig:tsne_stda_vs_mtda}, shows that our generalized approach for MTDA separates the features better than our single teacher. Regarding comparisons with the state-of-the-art STDA model with 1 model/target, our MTDA model is capable of reaching almost the same average performance while encoding multiple targets domains inside the same model.

\begin{table}[t!]
\centering
\caption{Accuracy of STDA baselines versus both versions of approach -- KD-STDA (single teacher on mixed targets) and KD-MTDA (multiple teachers with multiple targets) on the Office31 dataset}
\label{tb:Office31_stda_vs_mtda}
\resizebox{0.6\textwidth}{!}{
\begin{tabular}{|l|r|r|r|r|}
	\hline
	& \multicolumn{4}{|c|}{\textbf{Source $\rightarrow$ Targets}}  \\ 
	\textbf{Models}      & \textbf{A $\xrightarrow{}$ D,W} & \textbf{D $\xrightarrow{}$ A,W} & \textbf{W  $\xrightarrow{}$ A,D} & \textbf{Average} \\ \hline \hline
	CAT STDA\cite{CAT_ICCV} (1 model/target) & 78.5 & 62.9 & \textbf{98.8} & 80.1 \\ \hline \hline
	GCAN STDA\cite{GCAN_CVPR} (1 model/target) & \textbf{79.5} & \textbf{63.7} & 98.4 & \textbf{80.5} \\ \hline \hline
	KD-STDA RevGrad & 75.3                            & 64.0                            & 67.0                             & 68.8             \\ \hline
	KD-MTDA              & \textbf{82.5}                & \textbf{74.9}                & \textbf{77.6}                & \textbf{78.3}    \\ \hline
\end{tabular}
}
\end{table}

\subsection{Complexity analysis:} 

While CNNs pruned using the TCP \cite{TCP} method require fewer parameters than our student models, Table \ref{tb:complexity_comp} indicates that our method achieve the same number of FLOPS on ResNet34, and with even fewer FLOPs on ResNet18. This means that while TCP prunes more parameters, it may not have much impact on the number of FLOPS since the pruned filters are ranked and pruned globally across the network, instead of being pruned at each layer. TCP prunes away filters that do not impact the FLOPS but can impact performance. Another important point of having more parameters is that, over-parametrization can help provide better generalization -- our student CNNs have a greater chance to provide better generalization than a pruned model with fewer parameters. 

\begin{table}[htbp]
\caption{Computational complexity of proposed methods and TCP}
\label{tb:complexity_comp}
\centering
\resizebox{0.5\textwidth}{!}{
\begin{tabular}{|l||c|c|c|}
\hline
\textbf{Models} & \textbf{no. operations}  & \multicolumn{2}{|c|}{\textbf{no. parameters (M)}}\\
                &   \textbf{(GFLOPS)}  & \textbf{Office31}  & \textbf{ImageClef} \\ \hline \hline
ResNet50            & 4.1   & 25.5  & 25.5  \\ \hline
TCP(12\% Pruned)    & 3.6   & 15.8  & 15.9  \\ \hline
ResNet34 & 3.6   & 21.7  & 21.7  \\ \hline
TCP(46\% Pruned)    & 2.2   & 10.6  & 10.9  \\ \hline
ResNet18            & 1.8   & 11.1  & 11.1  \\ \hline
\end{tabular}
}
\end{table}

As for the complexity comparison of an STDA based approach adatped to MTDA setting, i.e. having one model for each target domain instead of having one model for all the target domains. If we assume that there are $N$ target domains and $M$ the memory space occupied by a backbone, this would meant that in the scenario of having one model per target domains, we would need $N \times M$ and in the worst case $N$ GPUs if it's a complex backbone. This clearly shows that having one model per target domain is not scalable and it is much preferable having one model to handle multiple target domains.

\section{Conclusion}

In this paper, we proposed a joint optimization of KD and UDA that tackles both the problem of domain shift and model compression in both the STDA and MTDA setting. In addition, our method also works with different UDA and KD techniques, whether it is logits or feature-based. Our results on STDA suggest that our method is capable of adapting and accelerating a model by alleviating the domain shift problem and reduce complexity. The proposed technique is also capable of adapting a model to multiple target domains while keeping high accuracy. In both settings, STDA and MTDA, our method outperforms the current state-of-the-art, especially on compact models. Since UDA is an active area of research, our future research can include a more efficient target domain knowledge transfer method for better compression or better fusion method for combining different target domains.

\section*{Acknowledgements}
This research was partially supported by the Mathematics of Information Technology and Complex Systems (MITACS) and the Natural Sciences and Engineering Research Council of Canada (NSERC) organizations.



\bibliography{mybibfile}

\newpage
\appendix
\clearpage
{ \centering \Large \textbf{Supplementary Material}}

\section{Hyper-parameters}

\begin{table*}[!htbp]
\centering
\caption{Hyper-parameters for our algorithms for each backbone and dataset}
\label{tb:hyper_parameters}
\resizebox{0.8\textwidth}{!}{
\begin{tabular}{|l|r|r|r|r|r|}
	\hline 
	\textbf{Hyper parameters}  & \textbf{Office31 Resnet34-18} & \textbf{ImageClef-DA ResNet34-18} & \textbf{Digits LeNet} & \textbf{Pacs LeNet} \\ \hline \hline
	$N_{e}$                                           & 400                       & 400                         & 100                        & 100                          \\ \hline
	$\tau$                                             & 20                        & 20                          & 20                         & 20                           \\ \hline
	$\alpha$                                          & 0.8                       & 0.8                         & 0.5                        & 0.5                          \\ \hline
	$s$                                              & 0.1                       & 0.1                         & 0.1                        & 0.1                          \\ \hline
	$f$                                               & 0.8                       & 0.8                         & 0.5                        & 0.5                          \\ \hline
	$\gamma$                                          & -                       & -                         & 0.5                        & 0.5                          \\ \hline
	UDA Learning Rate                              & 0.001                     & 0.0001                      & 0.01                      & 0.01                       \\ \hline
	KD Learning Rate                               & 0.001                      & 0.001                       & 0.01                       & 0.01                        \\ \hline
	weight decay                                  & 0.0005                    & 0.0005                      & 0.0005                     & 0.0005                       \\ \hline
\end{tabular}
}
\end{table*}

From Table \ref{tb:hyper_parameters}, we can find the details of our hyper-parameters for both STDA and MTDA setting on different dataset. These hyper-parameters were selected using a standard cross-validation process.

\section{Additional results}

\subsection{Logits distillation vs. feature distillation for MTDA} In this study, we compare the difference between logits distillation and feature distillation in MTDA setting on Office31 dataset. The Table \ref{tb:Office31_mtda_logits_vs_feature} shows the accuracy of our MTDA method with either logits or feature distillation on student model.

\begin{table}[ht]
\centering
\caption{Accuracy of proposed method using either logits or feature distillation}
\label{tb:Office31_mtda_logits_vs_feature}
\resizebox{0.8\textwidth}{!}{
\begin{tabular}{|l|l|l|l|l|}
\hline
& \multicolumn{4}{|c|}{\textbf{Source $\rightarrow$ Targets}} \\
\textbf{Types of distillation}                            & \textbf{A $\rightarrow$ D,W} & \textbf{D $\rightarrow$ A, W} & \textbf{W $\rightarrow$ A, D} & \textbf{Average} \\ \hline \hline
\multicolumn{5}{|c|}{\textbf{Teacher: ResNet50 --- Student: ResNet18}}                                                                                               \\ \hline
KD-MTDA logits distillation   & 80.8                & 79.7                 & 78.6                 & 79.7    \\ \hline
KD-MTDA feature distillation  & \textbf{81.2}                & \textbf{80.5}                 & \textbf{79.5}                 & \textbf{80.4}    \\ \hline
\end{tabular}
}
\end{table}

From Table \ref{tb:Office31_mtda_logits_vs_feature}, where we compare the performance of logits-based and feature-based distillation on the setting of MTDA, in contrast to with feature distillation for STDA, feature distillation in MTDA does not provide a significant improvement in performance. Results indicate that feature distillation in MTDA only perform slightly better than logits distillation. 

\section{t-SNE}

\begin{figure}[h!]
    \centering
    \includegraphics[width=0.8\textwidth]{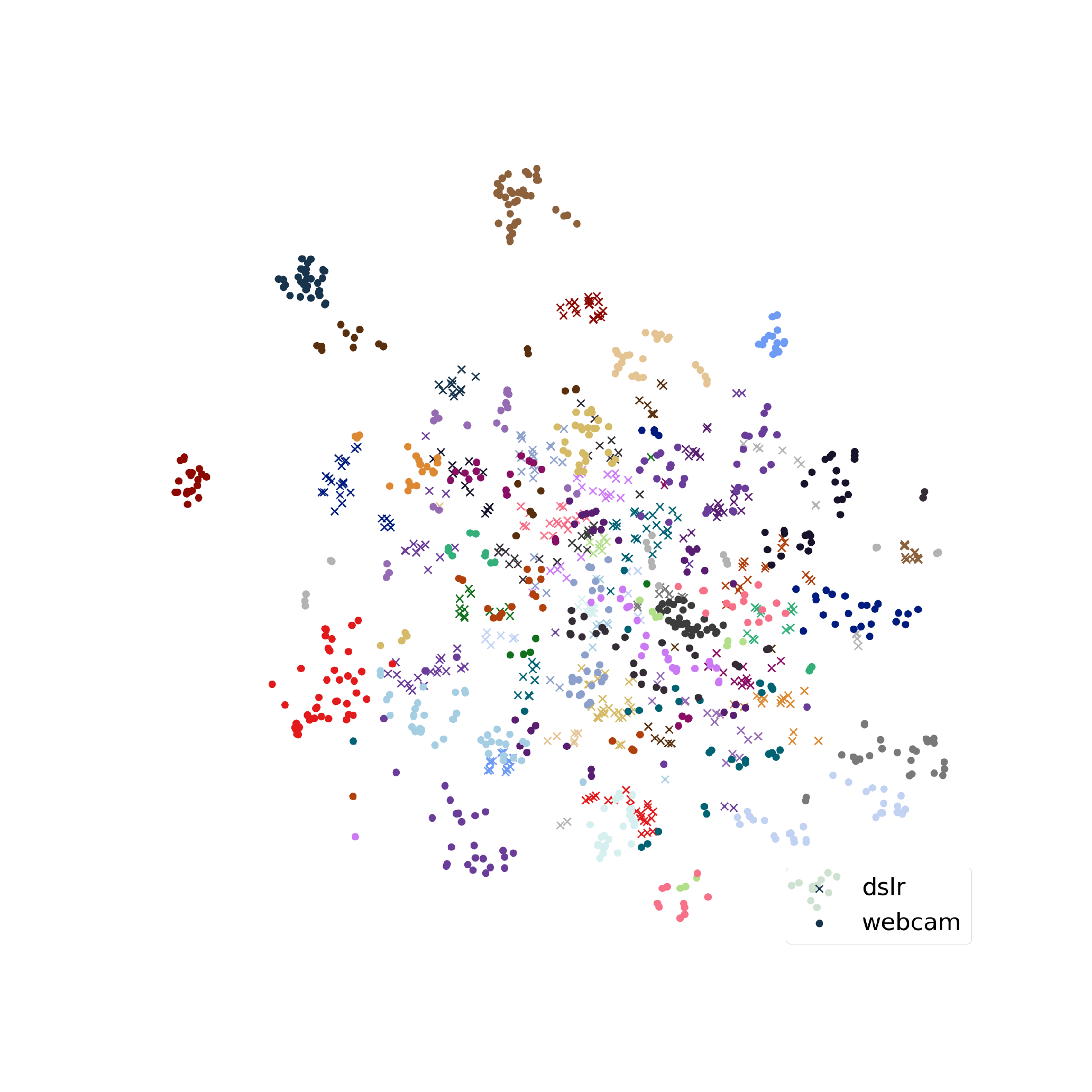}
    \caption{t-SNE of KD-MTDA. Best viewed in color.}
    \label{fig:tsne_kd_mtda}
\end{figure}

\begin{figure}[h!]
    \centering
    \includegraphics[width=0.8\textwidth]{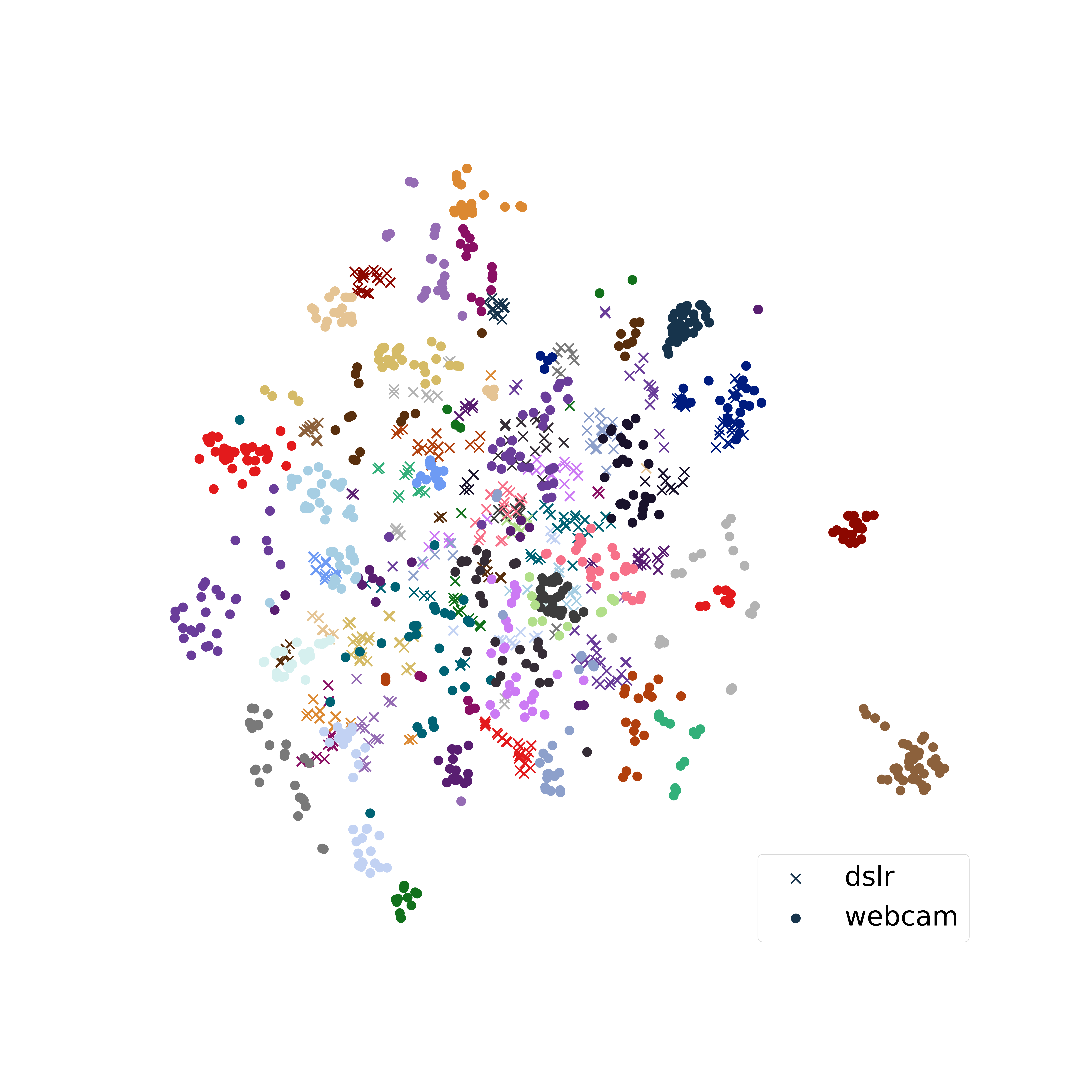}
    \caption{t-SNE of KD-STDA with mixed targets. Best viewed in color.}
    \label{fig:tsne_kd_stda}
\end{figure}

\begin{figure}[h!]
    \centering
    \includegraphics[width=0.8\textwidth]{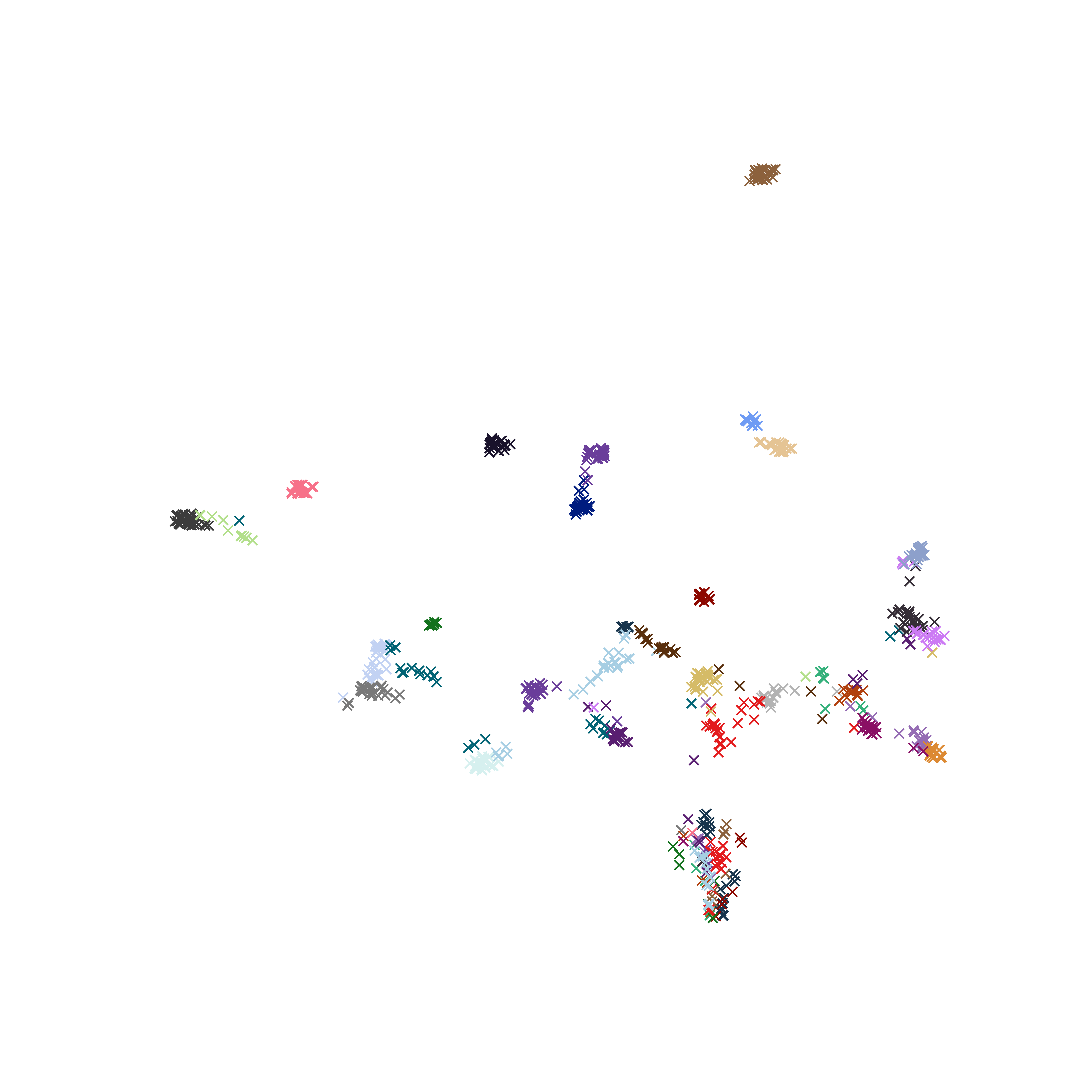}
    \caption{t-SNE of KD-STDA MMD. Best viewed in color.}
    \label{fig:tsne_mmd_stda}
\end{figure}

\begin{figure}[h!]
    \centering
    \includegraphics[width=0.8\textwidth]{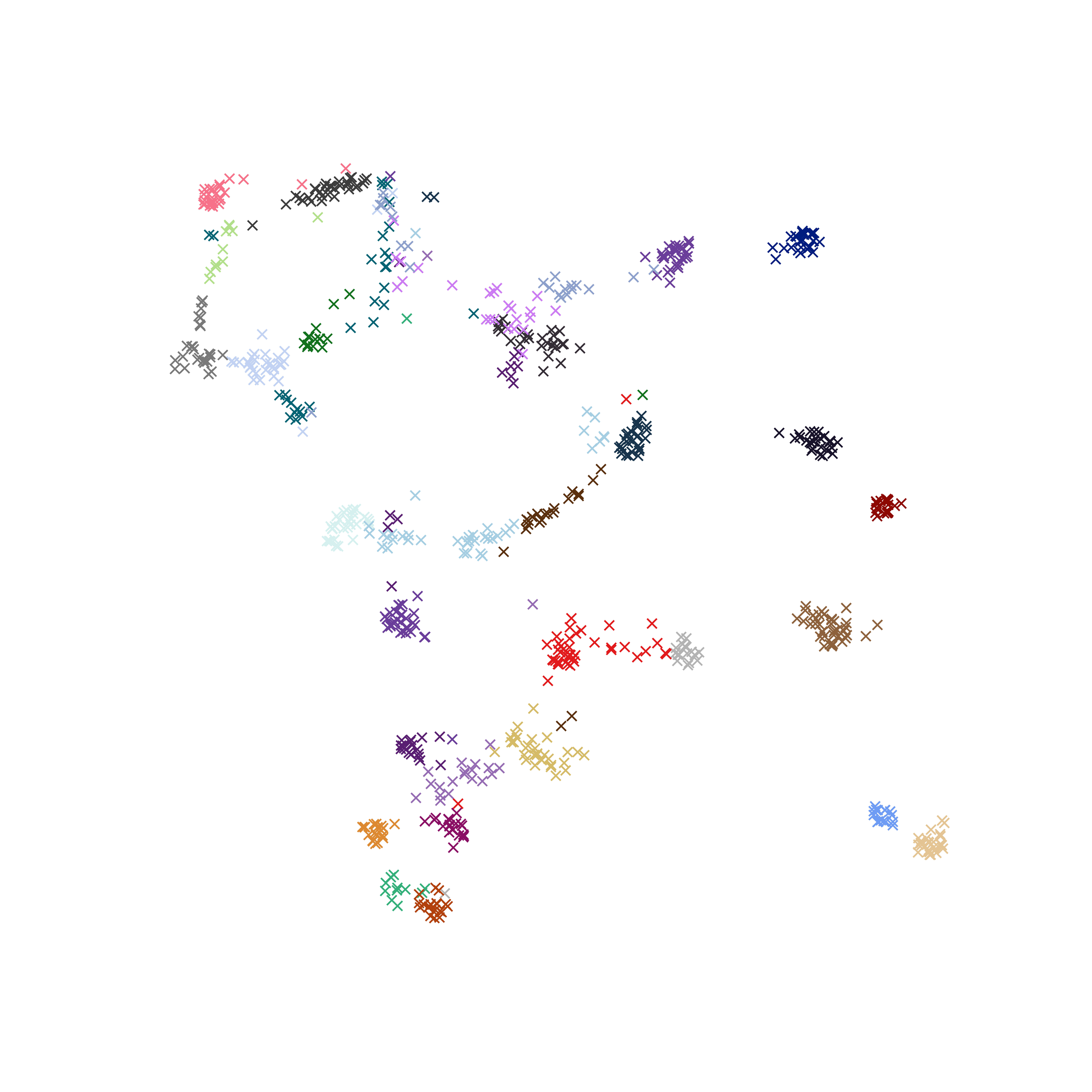}
    \caption{t-SNE of KD-STDA GRL. Best viewed in color.}
    \label{fig:tsne_grl_stda}
\end{figure}

\end{document}